\documentclass[twoside,11pt]{article}

\usepackage[final]{melba}  
\usepackage{booktabs}  
\usepackage{wrapfig}
\usepackage{caption}
\usepackage{comment}
\usepackage[T1]{fontenc}
\usepackage{xcolor}
\usepackage{amsmath,amsfonts}

\def \Rm{{\mathbb{R}}}

\def \xbf{{\mathbf x}}

\def \0bf{{\mathbf 0}}

\melbaheading{5}{https://www.melba-journal.org/article/21663-locally-orderless-tensor-networks-for-classifying-two-and-three-dimensional-medical-images?auth_token=HgMd7jGPhvS8EqDEmj30}{2021}{1-21}{09/2020}{03/2021}{Raghavendra Selvan and Silas Ørting and Erik B Dam}{Medical Imaging with Deep Learning (MIDL) 2020}{Marleen de Bruijne, Tal Arbel, Ismail Ben Ayed, Hervé Lombaert}

\ShortHeadings{LoTeNet for classification of 2D/3D medical images}{Selvan, Ørting and Dam}
\firstpageno{1}

\title{Locally orderless tensor networks for classifying two- and three-dimensional medical images}
\author{\name{Raghavendra Selvan} $^{1,2}$ \email{raghav@di.ku.dk}
\AND
\name{Silas Ørting} $^{1}$ \email{silas@di.ku.dk}
\AND
\name{Erik B Dam} $^{1}$ \email{erikdam@di.ku.dk}\\
\addr $^{1}$ Department of Computer Science, University of Copenhagen, Denmark\\
\addr $^{2}$ Department of Neuroscience, University of Copenhagen, Denmark
}

\begin{document}

\maketitle
\begin{abstract}

Tensor networks are factorisations of high rank tensors into networks of lower rank tensors and have primarily been used to analyse quantum many-body problems.  Tensor networks have seen a recent surge of interest in relation to supervised learning tasks with a focus on image classification. In this work, we improve upon the matrix product state (MPS) tensor networks
    that can operate on one-dimensional vectors to be useful for working with 2D and 3D medical images. We treat small image regions as orderless, {\em squeeze} their spatial information into feature dimensions and then perform MPS operations on these locally orderless regions. These local representations are then aggregated in a hierarchical manner to retain global structure. The proposed locally orderless tensor network (LoTeNet\footnote{{Source code is available here: \url{https://github.com/raghavian/LoTeNet_pytorch/}}}) is compared with relevant methods on three datasets. The architecture of LoTeNet is fixed in all experiments and we show it requires lesser computational resources to attain performance on par or superior to the compared methods.

\end{abstract}


\section{Introduction}

\begin{wrapfigure}{r}{0.5\textwidth}
    \centering
    \includegraphics[width=0.5\textwidth]{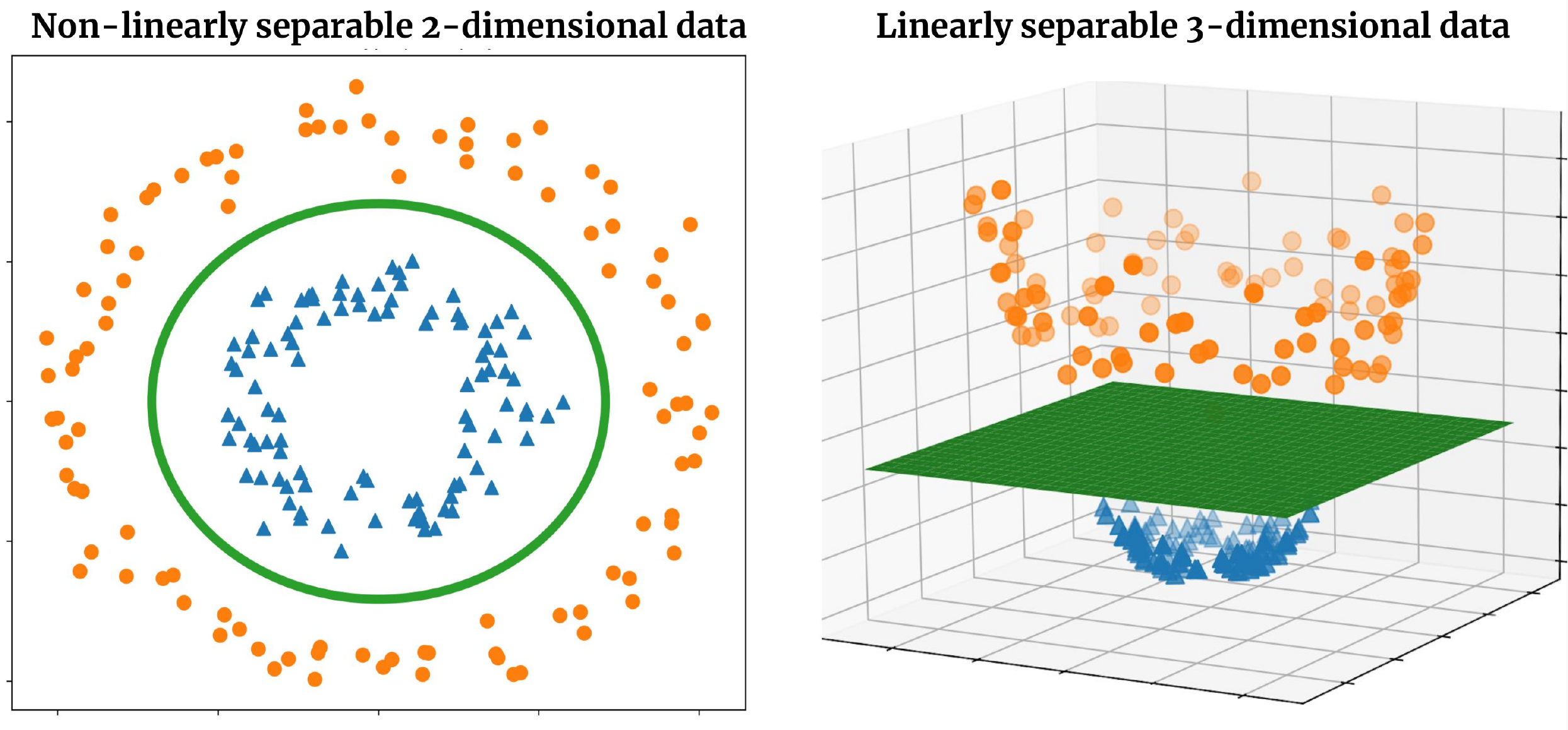}
    \caption{Data that is only non-linearly separable in lower dimensions can become linearly separable in higher dimensions, illustrated here for a simple case of 2-dimensional data that becomes linearly separable when lifted into 3-dimensions. This is the underlying principle in kernel based methods and tensor networks which learn linear decision boundaries in high dimensional spaces.}
    \label{fig:decision}
\end{wrapfigure}

Support vector machines (SVMs) and other kernel based methods ushered in a new wave of supervised learning methods based on the fundamental insight that challenging tasks in low dimensions may become easier when the data is lifted to higher dimensional spaces~\citep{boser1992training, cortes1995support, hofmann2008kernel}, as illustrated in Figure~\ref{fig:decision}. These methods, however, become prohibitive when dealing with massive datasets in high dimensional feature spaces as their space complexity grows at least quadratically with the number of data points~\citep{bordes2005fast, novikov2016exponential}. Further, SVMs are also known to be sensitive to the specific choice of the kernel parameters  and as a consequence the decision boundaries learnt are prone to over-fitting~\citep{burges1998tutorial,bordes2005fast}. These factors have discouraged their successful adoption to tasks involving large datasets comprising high resolution images where deep learning based methods have shown to fare better~\citep{litjens2017survey,liu2017svm}. 



 Tensor networks, also known as tensor trains, offer a different and more efficient framework to dealing with such high dimensional spaces. Fundamentally, tensor networks are factorisations of higher order~\footnote{Order of tensors are also referred to as ranks. In this work we adhere to using {\em order}.} tensors into networks of lower order tensors~\citep{fannes1992finitely,oseledets2011tensor,bridgeman2017hand}. The number of parameters needed to specify an order-$N$ tensor using tensor networks can be drastically reduced, from  {\em exponential to polynomial} dependence on $N$~\citep{perez2006matrix}. This massive reduction in number of parameters using tensor networks has been predominantly applied to better understand quantum wave functions~\citep{shi2006classical}. They have also seen applications in data compression~\citep{cichocki2016tensor}, and recently to better understand the expressive power of deep learning models~\citep{cohen2016expressive,glasser2019expressive}. 

 Recently, there has been an increasing interest in using tensor networks for supervised learning, specifically focused on image classification tasks~\citep{stoudenmire2016supervised,klus2019tensor,efthymiou2019tensornetwork,sun2020generative}. These methods rely on transforming two dimensional input images (order-2 tensors) into one dimensional vectors (order-1 tensors) to obtain linear decision boundaries in high dimensional spaces. Due to the constraint of flattening to obtain vector inputs these methods are constrained to work with images of small spatial resolution ($12$x$12$ px to $28$x$28$ px). Several improved flattening strategies have been attempted to maximize the retained spatial correlation between pixels as the correlation declines exponentially causing loss of information in larger images~\citep{stoudenmire2016supervised,efthymiou2019tensornetwork,cheng2019tree,sagan2012space}. For small enough images (like in MNIST~\footnote{\url{http://yann.lecun.com/exdb/mnist/}} or Fashion MNIST~\footnote{\url{https://github.com/zalandoresearch/fashion-mnist}} datasets) there is some residual correlation in the flattened images which can be exploited by lifting the vector input to higher dimensions using tensor networks. This might not be effective in the case of images with higher spatial resolution. Further, the information lost by flattening of images in medical imaging tasks can be critical as the predicted decisions could be dependent on the global structure of the image.

    In this work, we present a tensor network based model adapted for classifying two- and three-dimensional medical images which can be of higher spatial resolution. {The method presented here relies on lifting small image (slice or volume) patches to higher dimensions, performing tensor network operations on them to obtain intermediate representations which are then hierarchically aggregated to predict the final classification decisions}. These small, local regions can be treated as being {\em locally orderless} drawing parallels to the classical theory of locally orderless images in~\citet{koenderink1999structure}. As with locally orderless image analysis, we propose to extract useful representations of small image regions in higher dimensions and aggregate them in a hierarchical manner resulting in our locally orderless tensor network (LoTeNet) model. The proposed LoTeNet model is used to learn decision functions in high dimensional spaces in a supervised learning set-up and is optimized end-to-end by backpropagating the error signal through the tensor network. {The LoTeNet model builds on an earlier work that adapted tensor networks for supervised machine learning in~\citet{stoudenmire2016supervised}, and also has similarities to the tensor network model in~\citet{efthymiou2019tensornetwork}. Specifically, as with both these models, the LoTeNet model uses the matrix product state (MPS) tensor networks~\citep{perez2006matrix} to approximate linear decisions in a supervised learning set-up. The proposed modifications -- comprising the use of the hierarchical aggregation of patch-level representations of image data using tensor networks -- yield a large computational advantage to LoTeNet making it amenable to be used in medical image analysis.} 
    
    The performance of the LoTeNet model is demonstrated using experiments on classifying three medical imaging datasets: 2D histopathology images in PCam dataset~\citep{veeling2018rotation}, 2D thoracic computed tomography (CT) slices from LIDC-IDRI dataset~\citep{armato2004lung} and 3D magnetric resonance imaging (MRI) from the OASIS dataset~\citep{marcus2007open}. These experiments show that the LoTeNet model fares comparably to relevant state-of-the-art deep learning methods requiring the tuning of a single model hyperparameter while utilising only a fraction of the graphics processing unit (GPU) memory when compared to their convolutional neural network (CNN) counterparts. The work presented here is an extension of an earlier version of the method which operated on two-dimensional data and is published as a conference publication~\citep{raghav2020tensor}. 
\\
The key contributions in this work are:

    1. A novel tensor network based model for classifying 2D and 3D medical image data
    
    2. Extending supervised learning with tensor networks to images of higher resolution
    
    3. Validation of the method on three public datasets and comparison to relevant methods
    
    4. Demonstrating competitive performance using little GPU resources
    
    5. Conceptual connections of tensor networks with deep neural networks \\
In the remainder of this manuscript we present the basics of tensor notation, tensor network fundamentals and formulate the supervised image classification task in Section~\ref{sec:tensors}. The components of the proposed LoTeNet model and the final model itself are presented in Section~\ref{sec:method}. The proposed model is discussed in relation to existing literature in Section~\ref{sec:related}. Experiments on the three datasets and results are presented in Section~\ref{sec:res}, along with the discussions and future research directions in Section~\ref{sec:disc}, and present overall conclusions from this work in Section~\ref{sec:conc}.

\section{Background and Problem Formulation}
\label{sec:tensors}
We introduce key concepts pertaining the use and optimisation of tensor networks in this section which will be put together to describe the proposed method in the following sections.

\begin{figure}[t]
\centering
{\includegraphics[width=0.7\linewidth]{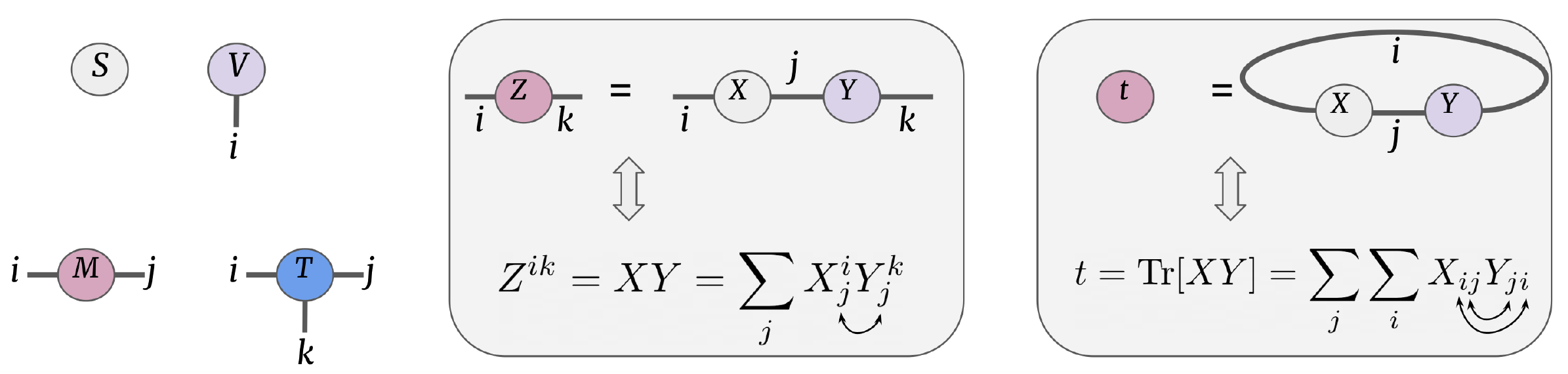}}
\caption{Left: Tensor notation depicting a scalar $S$, vector ${V}^i$, matrix $M^{ij}$ and a general order-$3$ tensor $T^{ijk}$. Center: Tensor notation for matrix multiplication or \emph{tensor contraction}, which are used extensively in the matrix product state networks used in this work. We adhere to the convention that the contracted indices are written as subscripts. Right: Tensor notation for trace of product of two matrices.}
\label{fig:tensorBasics}
\end{figure}

\subsection{Tensor Network Notations}

Tensor networks and operations on them are described using an intuitive graphical notation, introduced in~\citet{penrose1971applications}. Figure~\ref{fig:tensorBasics} (left) shows the commonly used notations for a scalar $S$, vector $V^i$, matrix $M^{ij}$ and a general order-$3$ tensor $T^{ijk}$. The number of dimensions of a tensor is captured by the number of edges emanating from the nodes denoted by the edge indices. For instance, the vector $V^i$ is an order-$1$ tensor indicated by the edge with index $i$ and an order-$3$ tensor has three indices $(i,j,k)$ depicted by the three edges, and so on. 

Tensor notations can also be used to capture operations on higher-order tensors succinctly as shown in Figure~\ref{fig:tensorBasics} (center) where matrix product is depicted, which is also known as \emph{tensor contraction}. The edge between the tensor nodes $X^i_j$ and $Y^k_j$ is the dimension subsumed in matrix multiplication resulting in the matrix $Z^{ik}$. 
More thorough introduction to tensor notations can be found in~\citet{bridgeman2017hand}.

\subsection{Linear Model in Exponentially High Dimensions}

A linear model in a sufficiently high dimensional space can be very powerful~\citep{novikov2016exponential}. In SVMs, this is accomplished by the \emph{implicit} mapping of the input data into an  infinite dimensional space using radial basis function kernels~\citep{hofmann2008kernel}. In this section, we describe the procedure followed in recent tensor network based works, including in ours, to map the input data into an exponentially high dimensional space.

Consider an input vector $\xbf \in [0,1]^N$, which can be obtained by flattening a 2D or 3D image with $N$ pixels with intensity values that are normalized in the interval $[0,1]$. A commonly used feature map for tensor networks is obtained from the \emph{tensor product} of pixel-wise feature maps~\citep{stoudenmire2016supervised}:
\begin{equation}
    \Phi^{i_1,i_2,\dots i_N}(\xbf) = \phi^{i_1}(x_1) \otimes \phi^{i_2}(x_2) \otimes \cdots \phi^{i_N}(x_N)
    \label{eq:jointRef}
\end{equation}
where the local feature map acting on a pixel $x_j$ is indicated by $\phi^{i_j}(\cdot)$. The indices $i_j$ run from $1$ to $d$ where $d$ is the dimension of the local feature map. The feature maps are usually simple non-linear functions restricted to have unit norm such that the joint feature map in Eq.~\eqref{eq:jointRef} also has unit norm. 
A widely used local feature map with $d=2$ inspired from quantum wave function analysis is~\citep{stoudenmire2016supervised} is shown in~\eqref{eq:localRef}, and a simpler local feature map from~\citet{efthymiou2019tensornetwork} in Eq.~\eqref{eq:localRef_1} with similar properties (but non-unit norm):
\begin{align}
    \phi^{i_j} (x_j) &= [\cos \left( \frac{\pi}{2}x_j\right), \sin \left(\frac{\pi}{2}x_j\right)]
    \label{eq:localRef} \\
    \phi^{i_j} (x_j) &= [x_j, 1-x_j].
    \label{eq:localRef_1}
\end{align}

The joint feature map $\Phi(\xbf)$ in Eq.~\eqref{eq:jointRef} is an order-$N$ tensor due to tensor product of the $N$  order-$1$ local feature maps of dimension $d$ in Eq.~\eqref{eq:localRef}. The joint feature map $\Phi(\xbf)$ can be seen as mapping each image to a vector in the $d^N$ dimensional feature space~\citep{stoudenmire2016supervised}. For multi-channel inputs,  such as RGB images or other imaging modalities, with $C$ input channels the local feature map can be applied to each channel separately such that the resulting space is of dimension $(d \cdot  C)^N$~\citep{efthymiou2019tensornetwork}. 

Given this high dimensional joint feature map $\Phi(\xbf)$ of Eq.~\eqref{eq:jointRef} for the input data $\xbf$, a linear decision rule for a multi-class classification task can be formulated as:
\begin{align}
    f(\xbf) &= \arg\max_m f^{m} (\xbf), 
    \label{eq:deRule} \\ 
f^{m}(\xbf) &= W ^{m} \cdot \Phi(\xbf).    
\label{eq:linModel}
\end{align}
where $ m=[0,1,\dots M-1]$ are the $M$ classes, and the weight tensor $W^{m}$ is  of order-($N+1$) with the output tensor index $m$. 

The tensor notation representation of the linear model of Eq.~\eqref{eq:linModel} is shown in Figure~\ref{fig:mps} (Step 1) where the first column of gray nodes are the individual pixel feature maps of order-$1$ with feature dimension $d$. The local feature maps are connected to the order-($N+1$) weight tensor $W^{m}$ along $N$ edges with one edge for output of dimension $M$ marked with index $m$. 

The order-($N$+1) weight tensor $W^{m}$ results in a total of $M \cdot d^N$ weight elements. Even for a relatively small gray scale image, say of size $100\times 100$, the total number of weight elements in $W^m$ can be massive: $2\cdot 2^{10000} \approx 10^{3010}$ which is exponentially more than the number of atoms in the universe\footnote{\url{https://en.wikipedia.org/wiki/Observable_universe}}! 
{
The $d^N$ lift in quantum physics is sometimes seen as a convenient illusion~\citep{poulin2011quantum,orus2014practical} as the most interesting behavior of systems can be captured with fewer degrees of freedom in this high dimensional space. This is analogous to using fewer than input feature dimensions after dimensionality reduction operations. In the next section we will see how tensor networks can access useful sub-spaces represented by such high dimensional tensors, with parameters that grow linearly with $N$ instead of growing exponentially with $N$.}

\subsection{Matrix Product State (MPS)}
\label{sec:mps}
\begin{figure}
    \centering
    \includegraphics[width=0.7\textwidth]{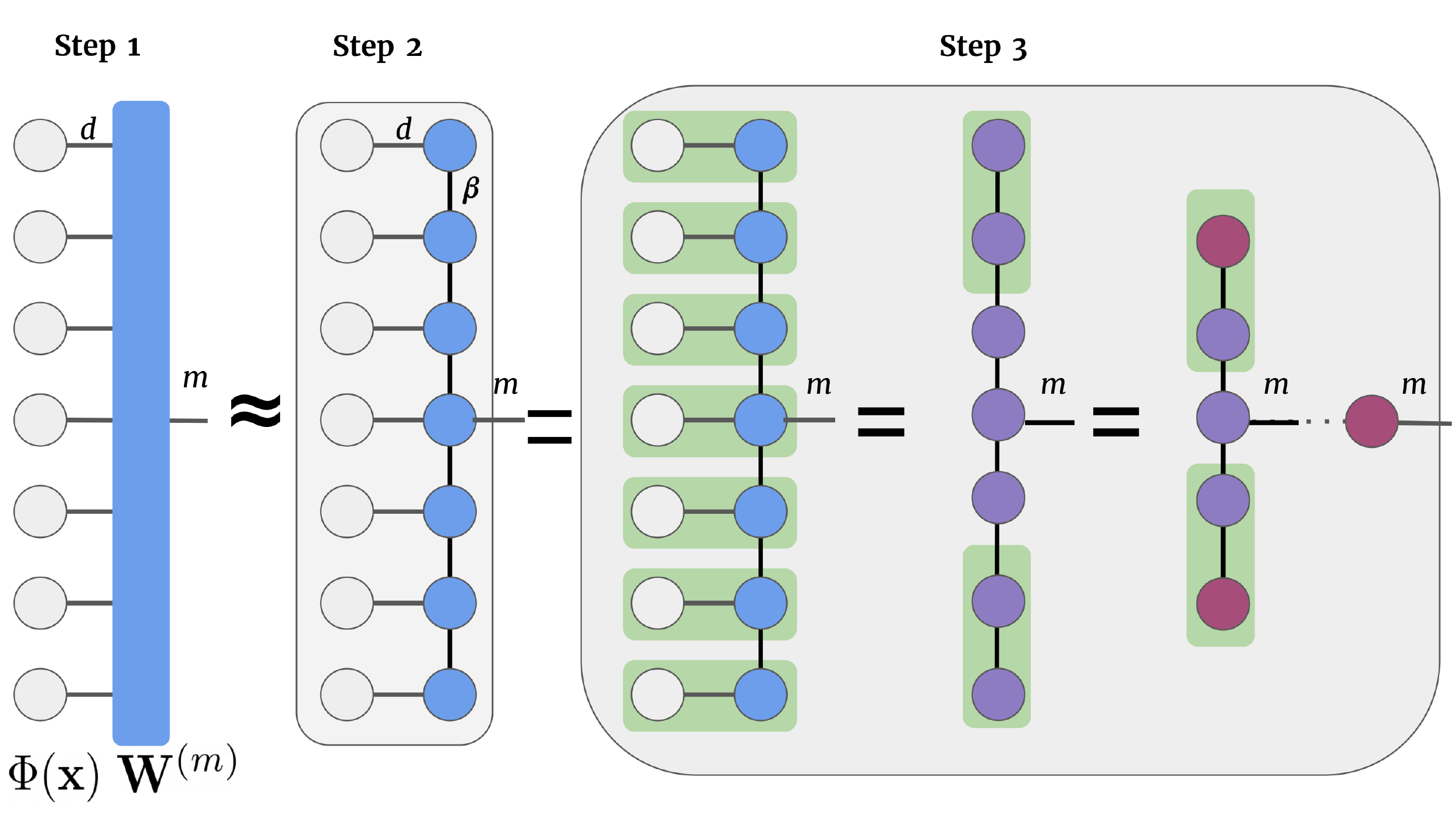}
    \caption{(Step 1) Linear model of Eq.~\eqref{eq:linModel} in tensor notation. (Step 2) MPS approximation of the linear model. (Step 3) Series of tensor contractions done with MPS to compute $W ^{m} \cdot \Phi(\xbf)$ in Eq.~\eqref{eq:linModel}}
    \label{fig:mps}
\end{figure}

Consider two vectors, $T^i$ and $U^j$  with indices $i$ and $j$ respectively. The tensor product~\footnote{Note that the tensor product of two order-1 tensors is the same as the vector outer product.} of these two vectors yields an order-$2$ tensor or a matrix $X^{ij}$. The matrix product state (MPS)~\citep{perez2006matrix} is a type of tensor network that expands on this notion of tensor products allowing the factorization of an order-$N$ tensor (with $N$ edges) into a chain of lower-order tensors. Specifically, the MPS tensor network can factorise an order-$N$ tensor with a chain of order-$3$ (with three edges) except on the borders where they are of order-$2$, as shown in Figure~\ref{fig:mps} (Step 2). Consider a tensor of order-$N$ with indices $i_i, i_2, \dots i_N$, using MPS it can be approximated using lower-order tensors as
\begin{equation}
    W^{m,i_1,i_2,\dots i_N} = \sum_{\alpha_1, \alpha_2,\dots \alpha_N} A^{i_1}_{\alpha_1} A^{i_2}_{\alpha_1 \alpha_2} A^{i_3}_{\alpha_2 \alpha_3} \dots A^{m,i_j}_{\alpha_j \alpha_{j+1}} \dots A^{i_N}_{\alpha_N},
    \label{eq:mps}
\end{equation}
where  $A^{i_j}$ are the lower-order tensors. The subscript indices $\alpha_j$ are virtual indices that are contracted and are of dimension $\beta$ referred as the \emph{bond dimension}. The components of these intermediate lower-order tensors $A^{i_j}$ form the tunable parameters of the MPS approximation. 
Note that any $N$ dimensional tensor can be represented exactly using an MPS tensor network if $\beta=d^{N/2}$. In most applications, however, $\beta$ is fixed to a small value or allowed to adapt dynamically when the MPS tensor network is used to approximate higher-order tensors~\citep{perez2006matrix,torchmps}.  
\begin{figure}
    \centering
    \includegraphics[width=0.6\textwidth]{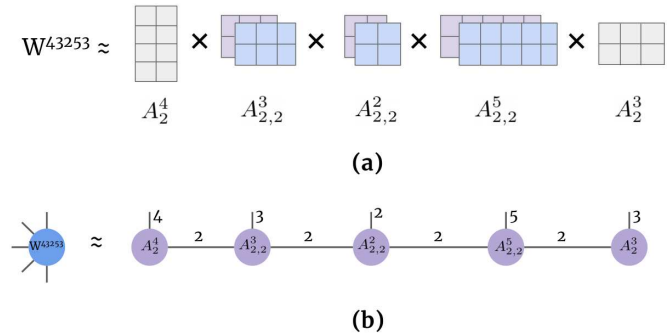}
    \caption{
{ {\bf (a)}: Illustration of MPS factorisation of an order-5 tensor $W^{43253}$ into five tensors of lower order (up to order-3 ) based on Equation~\eqref{eq:mps}. The bond dimension in this factorisation is $\beta=2$ seen as the subscript indices which are contracted. The tensor $W^{43253}$ has $4\times3\times2\times5\times3=360$ parameters whereas the MPS approximation requires  $8+12+8+20+6 = 54$ parameters. {\bf (b)}: MPS approximation of $W^{43253}$ in tensor notation.}}
    \label{fig:my_label}
\end{figure}

One of the drawbacks of using MPS tensor networks is that they operate along one dimension (as a chain). This is the primary reason that two dimensional image data has to be first flattened to a vector when working with tensor networks such as the MPS. Tensor networks that can work on arbitrary graphs,  which might be more suitable for image data like the projected entangled pair states (PEPS)~\citep{verstraete2004renormalization}, are not as well understood and do not yet have efficient algorithms like the MPS. 
\\
{\bf Dot product with MPS:} The decision function in Eq.~\eqref{eq:linModel} comprising the dot product of the order-($N$+1) weight tensor $W^m$ and the joint feature map $\Phi(\xbf)$ in Eq.~\eqref{eq:jointRef} can be efficiently computed using the MPS approximation in Eq.~\eqref{eq:mps}, depicted in Figure~\ref{fig:mps} (Step 2). The order in which tensor contractions are performed can yield a computationally efficient algorithm. The original MPS algorithm~\citep{perez2006matrix} starts from one of the ends, contracts a pair of tensors to obtain a new tensor which is then contracted with the next tensor and this process is repeated until the output tensor is reached. The cost of this algorithm scales with \{$N \cdot \beta^3 \cdot d$\} when compared to the cost without the MPS approximation which scales with $d^N$. In this work, we use the MPS implementation in~\citet{torchmps} which performs parallel contraction of the horizontal edges and then proceeds to contract them vertically as depicted in Figure~\ref{fig:mps} (Step 3). The MPS approximation also reduces the number of tunable parameters by an exponential factor, from \{$M \cdot d^N$\} to \{$M \cdot d \cdot N \cdot \beta^2$\}.

\section{Methods}
\label{sec:method}
The primary contribution in this work is a tensor network based image classification model that can handle high resolution images of both two and three spatial dimensions. Several recent tensor network models for supervised image classification tasks operate on small planar images and flatten them into vectors with different raveling strategies at the expense of global image structure~\citep{stoudenmire2016supervised,han2018unsupervised,efthymiou2019tensornetwork}. In contrast to these methods, we only flatten small regions of the images which can be assumed to be locally orderless~\citep{koenderink1999structure} and obtain expressive representations of local image regions using tensor contractions in high dimensional spaces. We process these locally orderless image regions in a hierarchical fashion using layers of MPS blocks in the final  model, which we call the locally orderless tensor network (LoTeNet), shown in Figure~\ref{fig:LoTeNet}. We next describe the proposed LoTeNet model in detail.

\subsection{Squeeze operation} 
\label{sec:squeeze}
\begin{wrapfigure}{r}{0.5\textwidth}
\centering
\includegraphics[width=0.45\textwidth]{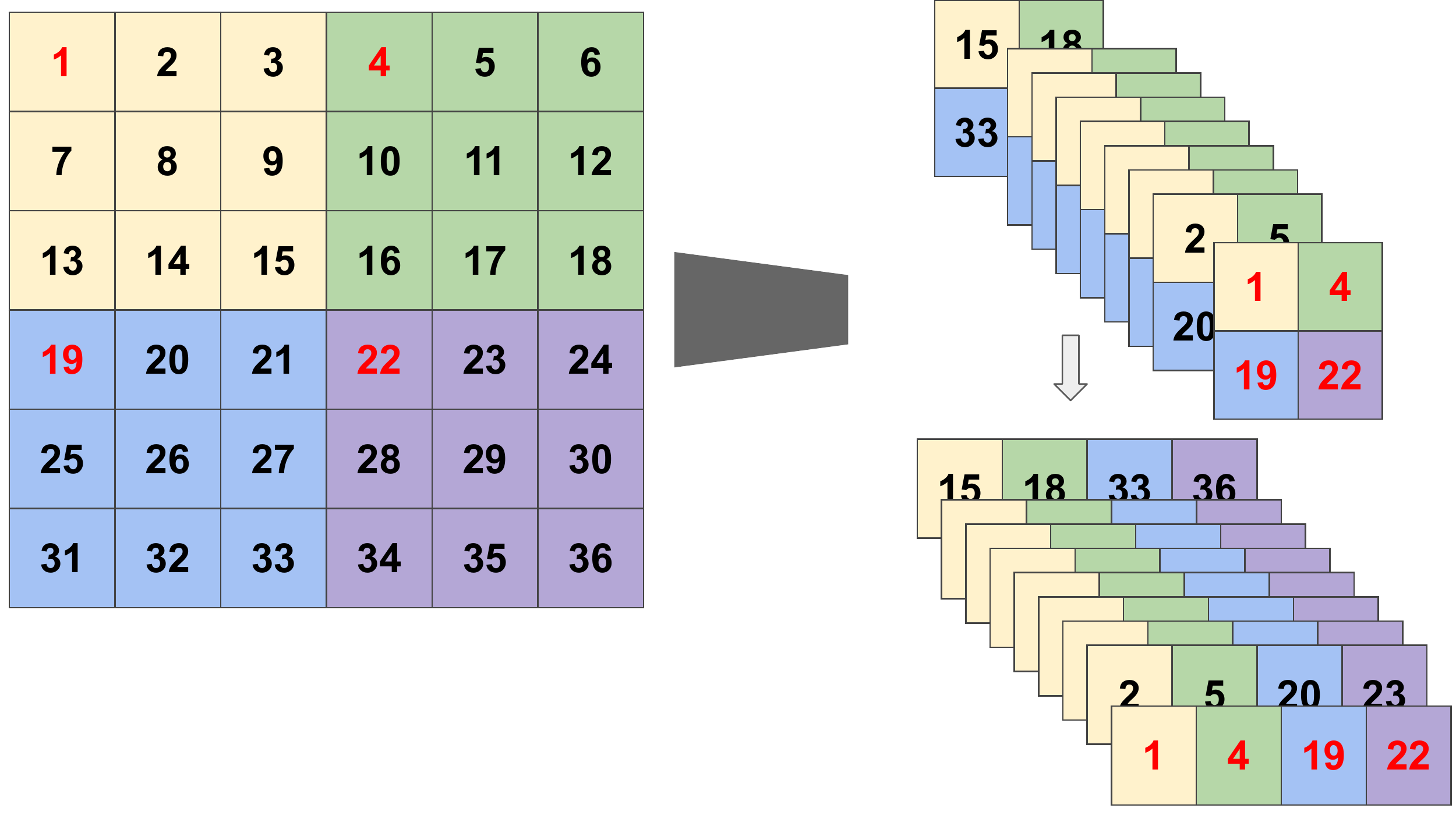}
\captionsetup{format=plain}
\caption{Squeeze operation with stride $k=3$  which reshapes a $6\times 6 \times 1$ image patch into a vector of size $4$ with feature dimension d=$9$.}
\label{fig:squeeze}
\end{wrapfigure}
The proposed LoTeNet model operates on small image regions and aggregates them with MPS operations at multiple levels. These small image patches at each level are created using the {\em squeeze} operation in two steps, as illustrated in Figure~\ref{fig:squeeze} for an order-$2$ tensor i.e an image with two spatial dimensions.

The input images are converted into vectors with inflated feature dimensions by applying kernels of stride $k$ along each of the spatial dimensions. Consider an input image $X^{hij}$ with $S=3$ spatial dimensions (height:H, width:V, depth:D), $N$ pixels and $C=d$ channels as feature dimension: $X^{hij} \in \Rm^{H \times V \times D \times C}$. In the first step, input image is reshaped into smaller patches controlled by the stride $k$:~$X^{hij} \in \Rm^{ (H/k) \times (V/k) \times (D/k) \times d}$ such that the feature dimension increases to $d=C \cdot k^S$. The stride of the kernel $k$ decides the extent of reduction in spatial dimensions and the corresponding increase in feature dimension of the squeezed image. 

In the second step, the reshaped image patches with inflated feature dimensions are flattened from order-$S$ tensors to order-$1$ tensors, resulting in $X^h \equiv \xbf \in \Rm^{(N/k^S) }$ with $d= C\cdot k^S$. This flattening operation provides spatial information in the feature dimension to LoTeNet, thus retaining additional image level structure when compared to flattening the entire image into a single vector $\xbf \in \Rm^{N}$ with $d=C$ channels~\citep{efthymiou2019tensornetwork}. Further, the increase in the feature dimension $d=C \cdot k^S$ makes the tensor network more expressive as it increases the dimensionality of the feature space~\citep{stoudenmire2016supervised}.

The transformations in dimensions due to the squeeze operation parameterised by the stride $k$, $\psi(\cdot;k)$, are summarised below:
\begin{equation}
\psi(\cdot;k): \{ X^{hij} \in \Rm^{H \times V \times D \times d}, d=C\}  \longrightarrow \{\xbf \in \Rm^{ (N/k^S) \times d}, d=C \cdot k^S\}.
\label{eq:squeeze}
\end{equation}
The squeeze operation can also be treated as a local feature map due to the increase in feature dimensions, similar to the more formal pixel level feature maps mentioned in Eq.~\eqref{eq:localRef} and Eq.~\eqref{eq:localRef_1}, operating on image patches instead of individual pixels. Finally note that the squeeze kernel stride can be different at each level $l$ of the model denoted $k_l$.

In the current formulation, the images are assumed to be rectangular in each plane. As the main purpose of the squeeze operation is to increase the feature dimensions by flattening small local neighbourhoods, this can be achieved by reshaping any small, non-square region of the image into a vector of fixed feature dimension.  The simplest strategy to work with non-square (circular images for instance) would be to pad the regions around the borders to make them rectangular. Another approach could be to partition non-square images into cells using Voronoi tesselations~\citep{du1999centroidal} and squeezing these cells into vectors. 


\subsection{Locally orderless tensor network (LoTeNet)}
\label{sec:LoTeNet}


\begin{figure}[t]
\centering
  {\includegraphics[width=0.89\linewidth]{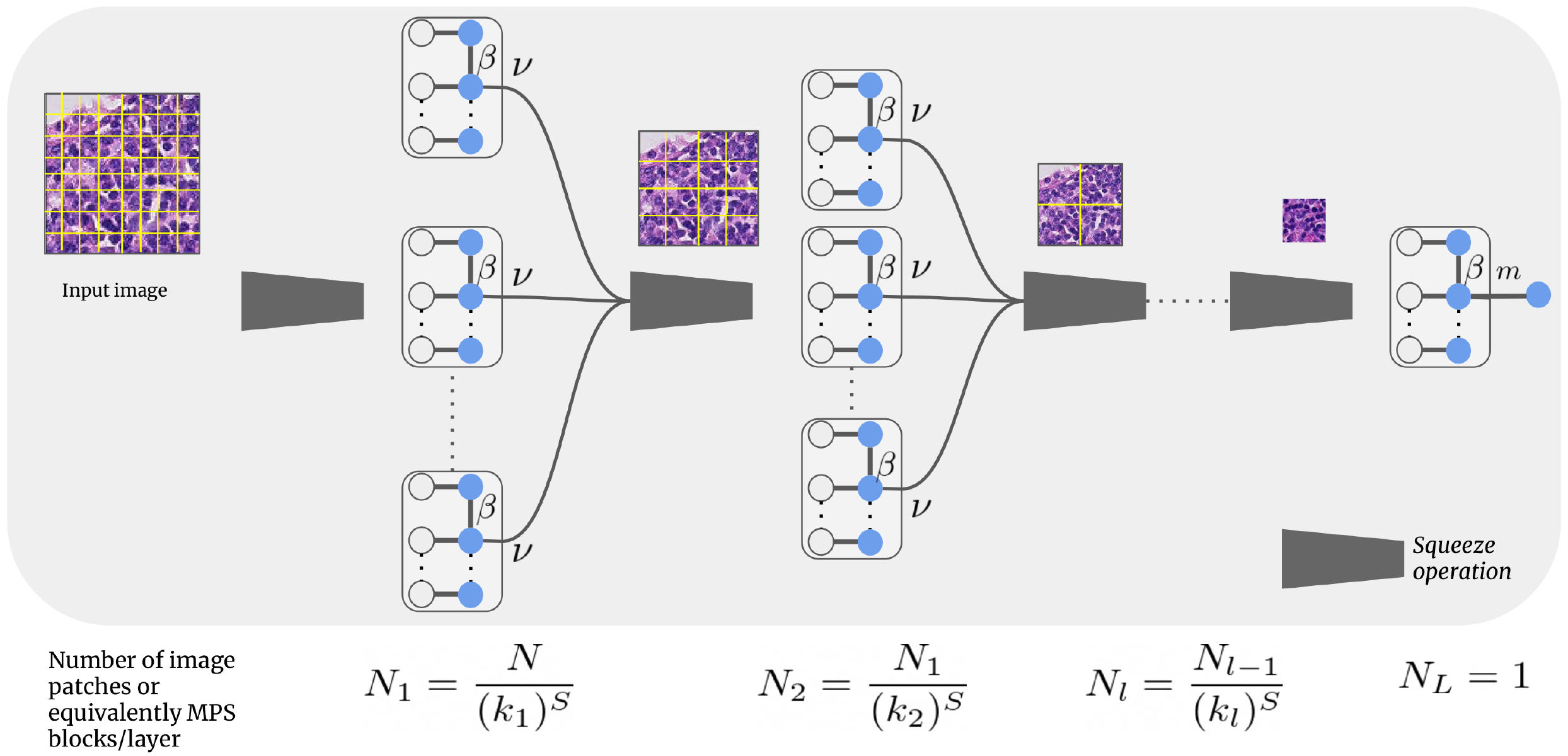}}
  \caption{The proposed locally orderless tensor network (LoTeNet) with $L$ layers. Each layer $l$ consists of several MPS blocks based on the squeeze kernel stride $k_l$ at that layer. The squeeze operation is as described in Figure~\ref{fig:squeeze}. The final MPS block outputs the $M$ class predictions shown as the edge with index $m$.}
  \label{fig:LoTeNet}
\end{figure}

An overview of the proposed LoTeNet model is shown in Figure~\ref{fig:LoTeNet} for an image with two spatial dimensions i.e with $S=2$. The number of squeezed vectors and the corresponding MPS blocks at each layer of the proposed model are also indicated in Figure~\ref{fig:LoTeNet}. 

The LoTeNet model comprises of layers of MPS blocks interleaved with squeeze operations {\em without any} non-linear components. The input to the first layer in Figure~\ref{fig:LoTeNet} is the full resolution input image with $N$ pixels, shown with grids marking the $k \times k$ patches which are then squeezed to obtain $N_1 = N/k^S$ vectors with $d=C\cdot k^S$. Each of these squeezed patches are input into MPS blocks which contract the $d=C\cdot k^S$ vectors to output a vector with dimension $d=\nu$. MPS blocks operating on squeezed image patches can be interpreted as summarising them with a vector of size $\nu$ using a linear model in a higher dimensional feature space. In LoTeNet, we set $\nu$ to be the same as the MPS bond dimension $\beta$, so that there is only a single hyperparameter to be tuned.

The output vectors from all $N_l$ MPS blocks at a given layer $l$ are reshaped back into the $S$ dimensional image space. However, due to the MPS contractions, these intermediate image space representations will be of lower resolution as indicated by the smaller image with fewer patches in Figure~\ref{fig:LoTeNet}. This is analogous to obtaining an average pooled version of the intermediate feature maps in traditional CNNs. The intermediate images of lower resolution formed at layer $l$ are further squeezed and contracted in the subsequent layers of the model. This process is continued for $L-1$ layers and the final MPS block acting as layer $L$ contracts the output vector from layer $L-1$ to predict the decisions in Eq.~\eqref{eq:linModel}. The weights of all the MPS blocks in each of the $L$ layers are the model parameters which are tuned in a supervised setting as described next. 

\subsection{Model Optimization}

We view the sequence of MPS contractions in successive layers of our model as forward propagation and rely on automatic differentiation to compute the backward computation graph~\citep{efthymiou2019tensornetwork}. Torch MPS package~\citep{torchmps} is used to implement MPS blocks and trained in PyTorch~\citep{paszke2019pytorch} to learn the model parameters from training data in an end-to-end fashion. 
These parameters are analogous to the weights of neural network layers and can be updated in a similar iterative manner by backpropagating a relevant error signal computed between the model predictions and the training labels.

We minimize the cross-entropy loss between the true label $y_i \in [0,\dots,M-1]$ for each image $\xbf_i \in \mathcal{D}$ and the predicted label $f^{(y_i)}(\xbf_i)$ in the training set $\mathcal{D}$:
\begin{equation}
    \mathcal{L}(f^{(y_i)}(\xbf_i)) = -\sum_{(\xbf_i,y_i)\in \mathcal{D}} \log \frac{\exp{f^{(y_i)}(\xbf_i)}}
    { \sum_{m=0}^{M-1} \exp{f^{(m)}(\xbf_i)}} 
    = - \sum_{(\xbf_i,y_i)\in \mathcal{D}} \log \Big({\sigma}(f^{(y_i)}(\xbf_i))\Big)
    \label{eq:loss}
\end{equation}
where $\sigma(\cdot)$ is the softmax operation used to obtain normalized scores that can be interpreted as the predicted class probabilities. For binary classes, the loss in Eq.~\eqref{eq:loss} reduces to the binary cross entropy and output dimension $M=1$ can be used with the sigmoid non-linearity to obtain probabilistic predictions in $[0,1]$.

{
\section{Related work}
\label{sec:related}
Key ideas in the proposed locally orderless tensor network are related to several existing methods in the literature. The closest of them is the tensor network model in~\citet{efthymiou2019tensornetwork} which in turn was based on the work in~\citet{stoudenmire2016supervised}. The primary difference of LoTeNet compared to~\citet{efthymiou2019tensornetwork} is in its ability to handle high resolution 2D/3D images without losing spatial correlation between pixels. This capability is incorporated with strategies from image analysis such as the interpretation of small regions to be orderless~\citep{koenderink1999structure}. Further, the multi-layered approach used in LoTeNet has similarities with the multi-scale analysis of images~\citep{lindeberg2007scale}.}

{The {\em squeeze} operation described in Section~\ref{sec:squeeze} serves dual purposes: to move spatial information into feature dimension and to increase the feature dimension. This is based on similar operations performed in normalizing flow literature to provide additional spatial context via feature dimensions such as in~\citet{dinh2016density}.  The operation of stacking pixels into feature dimension is also similar to the {\em im2col} operations used to transform convolution into multiplications for improving efficiency of deep learning operations~\citep{chetlur2014cudnn}. Recently, in~\citet{blendowski2020learning}, the {\em im2col} operation was also used as a preprocessing step with differentiable decision trees such as random ferns~\citep{ozuysal2009fast}. The squeeze operation can also been as the inverse of the pixel shuffle operation introduced in~\citet{shi2016real} where the feature maps are interleaved into spatial dimension to perform sub-pixel convolution.}
\section{Data and Experiments}
\label{sec:res}

\subsection{Data}

\begin{figure}[t]
\centering
  {\includegraphics[width=0.95\linewidth]{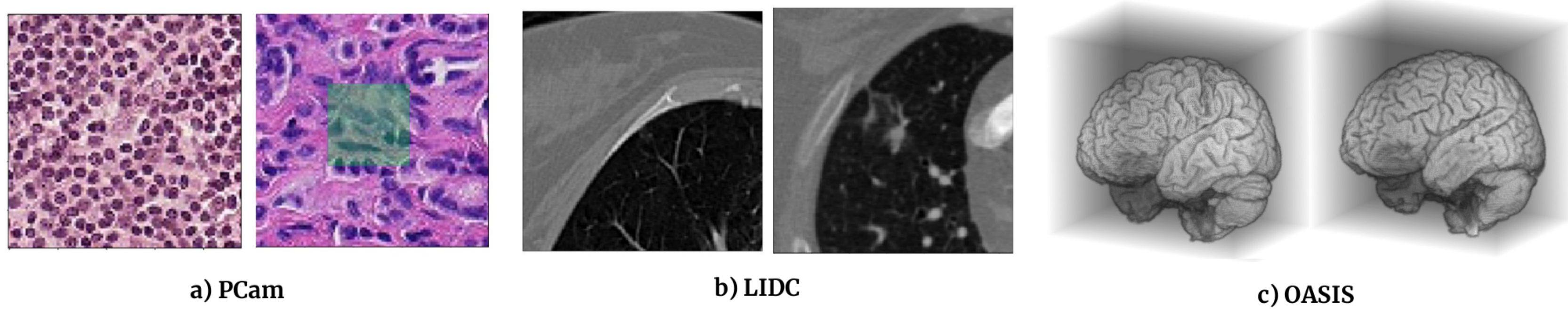}}
  \caption{Two sample images from each of the three datasets a) PCam~\citep{veeling2018rotation}, b) LIDC~\citep{armato2004lung} and c) OASIS~\citep{marcus2007open}.}
  \label{fig:dataset}
\end{figure}


We report experiments on three public datasets for the task of binary classification from 2D histopathology slides, 2D thoracic computed tomography (CT) and 3D T1-weighted MRI scans. 
\\
%
\textbf{PCam Dataset}: The PatchCamelyon (PCam) dataset is a binary histopathology image classification dataset introduced in~\citet{veeling2018rotation}. Image patches of size $96\times 96$ px are extracted from the Camelyon16 Challenge dataset~\citep{bejnordi2017diagnostic} with positive label indicating the presence of at least one pixel of tumour tissue in the central $32\times 32$ px region and a negative label indicating absence of tumour, as shown in Figure~\ref{fig:dataset}(a). In this work, a modified PCam dataset from the Kaggle challenge\footnote{\url{https://www.kaggle.com/c/histopathologic-cancer-detection}} is used that removes duplicate image patches included in the original dataset. This dataset consists of about $220,000$ patches for training and an independent test set of about $57,500$ patches is provided for evaluating the models. Data augmentation is performed during training with $0.5$ probability comprising of rotation, horizontal and vertical flips.\\
%
\textbf{LIDC Dataset}: The LIDC-IDRI datatset comprises of 1018 thoracic CT images with lesions annotated by four radiologists~\citep{armato2004lung}. We use the $128 \times 128$ px 2D slices from~\citet{lidc} (shown in Figure~\ref{fig:dataset} (b)) yielding a total of $15,096$ patches. The segmentation masks from the four raters are transformed to binary labels indicating the presence (if two or more radiologists marked a tumour) or absence of tumours (if less than two raters marked tumours). These binary labels capture a majority voting among the radiologists. The dataset is split into $60:20:20$ splits for training, validation and a hold-out test set. 
\\
\textbf{OASIS Dataset}: The OASIS dataset consists of T1-weighted MRI scans of 416 subjects aged between 18 to 96 years~\citep{marcus2007open}. The scans are graded with clinical dementia rating (CDR) into four classes: non-demented=0, very mild dementia=0.5, mild dementia=1.0, moderate dementia=2.0. We create an Alzheimer's disease classification dataset according to~\citet{wen2020convolutional} with scans of all subjects above 60 years resulting in a dataset with 155 subjects. Binary labels are extracted from the CDR scores: cognitively normal (CN) if CDR=0 and Alzheimer's disease (AD) if CDR$>$0, yielding 82:73 split between CN and AD cases. The MRI scans are preprocessed using the {\em extensive} pipeline described in~\citet{wen2020convolutional} comprising bias field correction, non-linear registration and skull stripping using the ClincaDL package\footnote{\url{https://clinicadl.readthedocs.io/en/latest/}}. After preprocessing we obtain volumes of size 128x128x128 voxels (Figure~\ref{fig:dataset}-c). 

\subsection{Experiments and Results}
\label{sec:exp}

\subsubsection{Implementation}
The model is implemented in PyTorch~\citep{paszke2019pytorch} based on the Torch MPS package~\citep{torchmps}. The architecture of LoTeNet model is kept fixed across all the datasets. The proposed LoTeNet model is evaluated with $L=4$ layers, squeeze kernel of size $k_l=2$ except for the input layer where $k_1=8$ in order to reduce the number of MPS blocks in the first layer.  The most critical hyperparameter of LoTeNet is its bond dimension $\beta$; it was set to $\beta=5$ obtained from the range $[2,3 \dots 20]$ based on the performance on the PCam validation set. We set the virtual dimension $\nu$ to be the same as the bond dimension. 
{Note that increasing the depth in LoTeNet does not translate to obtaining more complex decisions (like in neural networks) as LoTeNet is a linear model. Number of layers are controlled by the kernel stride and primarily result in reduced computation cost~\citep{selvan2020multi}. Finally, to keep the number of tunable hyperparameters lower, we also do not optimize $\beta$ and $L$ jointly which would yield a model similar to an adaptive MPS~\citep{stoudenmire2016supervised,torchmps}}

We use the Adam optimizer~\citep{kingma2014adam} with a learning rate of $5\times 10^{-4}$. We use a batch size of $512$ for experiments on the PCam and LIDC datasets, and a batch size of $4$ for the OASIS dataset. We incorporate batch normalisation~\citep{ioffe2015batch} after each layer in all the models and we observed this resulted in faster and more robust convergence. All models were assumed to have converged if there was no improvement in validation accuracy over $10$ consecutive epochs and the model with the best validation performance was used to predict on the test set. All experiments were run on a single Tesla K80 GPU with 12 GB memory. The development and training of all models (including baselines) in this work is estimated to produce 14.1 kg of CO2eq, equivalent to 120.1 km travelled by car as measured by Carbontracker~\footnote{\url{https://github.com/lfwa/carbontracker/}}~\citep{anthony2020carbontracker}.

\subsubsection{Experiments on 2D data} 
\begin{table}[t]
\small
\centering
    \caption{Performance comparison on PCam dataset (left) and LIDC dataset (right). For all models, we specify the GPU memory utilisation in gigabytes. Best AUC across all models is shown in boldface.}
    \label{tab:pcam}
    \begin{minipage}{.39\linewidth}
     \centering
    \begin{tabular}{lccc}
    \toprule
    PCam Models & GPU(GB) & AUC  \\
    \midrule
    Rotation Eq-CNN & $11.0$ &${\bf 0.963}$     \\
    {Densenet} &$10.5$& $0.962$    \\
    {LoTeNet (ours)} & $0.8$ &$0.943$ \\
    {Tensor Net-X ($\beta=10$)} & $5.2$ & $0.908$ \\
    \bottomrule
  \end{tabular}
    \end{minipage}%
    \hspace{0.65cm}
    \begin{minipage}{.55\linewidth}
      \centering
  \begin{tabular}{lccc}
    \toprule
LIDC Models & GPU(GB)& AUC\\
    \midrule
    {LoTeNet (ours)} & $0.7$ & ${\bf 0.874}$  \\
    {Tensor Net-X ($\beta=10)$} & $4.5$ & $0.847$ \\
    {Densenet} & $10.5$ & $0.829$ \\
    {Tensor Net-X ($\beta=5)$} & $1.5$ & $0.823$ \\
    \bottomrule
  \end{tabular}
    \end{minipage}
\end{table}

{\bf PCam models:} Performance of the LoTeNet model is compared to the rotation equivariant CNNs in~\citep{veeling2018rotation} which is also state-of-the-art for the PCam dataset. Additionally, we compare to the single layer MPS model with local feature map of the form in Eq.~\eqref{eq:localRef_1} with $\beta=10$ from~\citet{efthymiou2019tensornetwork} reported here as Tensor Net-X and to Densenet~\citep{huang2017densely} with $4$ layers and a growth rate of $12$. We compare the binary classification performance of the models using the area under the ROC curve (AUC) as it is not sensitive to arbitrary decision thresholds. The performance metrics are shown in Table~\ref{tab:pcam} (left) along with the maximum GPU memory utilisation for each of the models. We observe the AUC on the test set attained by LoTeNet model ($0.943$) is comparable to the Rotation Eq. CNN ($0.963$) and Densenet models ($0.962$) on the PCam dataset with a drastic reduction in the maximum GPU memory utilisation; a mere 0.8 GB when compared to upto 11 GB for the Rotation Eq.CNN~\footnote{In~\citet{veeling2018rotation} the authors used 4x12GB GPUs which was confirmed to R.Selvan by B.Veeling.} and 10.5 GB for Densenet. We also notice a considerable improvement when compared to Tensor Net-X in the attained AUC (0.908). 
\\
{\bf LIDC models:}
Similar to the PCam dataset, performance of LoTeNet model is compared using AUC and maximum GPU memory utilisation, and are reported in Table~\ref{tab:pcam} (right). We compare to Densenet and Tensor Net-X with two bond dimensions ($\beta=5,10$) to highlight the influence of the bond dimension. LoTeNet model fares better than the compared models with an AUC of $0.874$ whereas the Densenet model achieves $0.829$ and Tensor Net-X achieves $0.847$ ($\beta=10$) and $0.823$ ($\beta=5$). The GPU memory utlisation follows a similar trend as with the PCam models with LoTeNet requiring only $0.7$ GB.


\subsubsection{Experiments on 3D data} 

LoTeNet used in the 3D experiments is identical to the one used in 2D experiments except for the local feature map. For the 3D experiments, we do not apply the local feature map in Eq.~\eqref{eq:localRef} but instead rely on the increase in feature dimension due to the squeeze operation in Eq.~\eqref{eq:squeeze}. This reduces the number of parameters of the LoTeNet model and the risk of overfitting to the limited data in the OASIS dataset. 

The experimental set-up closely follows from~\citet{wen2020convolutional} and we use 5-fold cross validation to report the balanced accuracy on each of the test folds. Balanced accuracy (BA) is binary accuracy normalised by the class skew and helps when the classes are imbalanced. We reimplemented the subject-level 3D CNN used in~\citet{wen2020convolutional} as one of the compared methods. The subject-level CNN comprises 5 layers of 3D CNN with an initial feature map of $8$, max pooling and rectified linear units (ReLU) with doubling of feature maps after each layer, and an additional three linear layers with ReLUs to output the final predictions. We also report two other baseline models based on 3D CNN and a multi-layered perceptron (MLP) model. The CNN baseline model is similar to the subject-level CNN but has an initial feature map size of $32$. The MLP baseline has four layers to match the LoTeNet model, and ReLU non-linear activation function. 

Table~\ref{tab:oasis} summarises the performance of LoTeNet along with the compared methods for two types of inputs. As LoTeNet is agnostic to the dimensionality of the input image due to the squeeze operations, we demonstrate the performance on volume data (3D) and also on 19,840 slices extracted along the z-axis (2D) from the volume data. For the 2D experiments we ensure there is no data leakage by creating folds at subject level. The balanced accuracy across five test folds are shown and we see a clear improvement ($0.71\pm 0.09$) for the 3D case compared to subject-level CNN model ($0.67 \pm 0.08$). LoTeNet model has a lower score in the 2D case, indicating the model is able to extract more global information from the volume data than from 2D slices. Additionally, we report the number of parameters, maximum GPU utilisation and time per training epoch for all the methods. To show the influence of the number of parameters, we compare to a large MLP baseline model with 78M parameters and observe this increase does not necessarily improve the baseline model's performance. 

\begin{table}[t]
\small
\centering
    \caption{Performance comparison on OASIS dataset with comparing methods using 3D and 2D inputs. The performance is reported as balanced accuracy (BA) averaged over 5-fold cross validation. Number of parameters, maximum GPU utilization (GPU) and computation time per training epoch (t) for all methods are also reported.}
    \label{tab:oasis}
  \begin{tabular}{lcrrrc}
    \toprule
OASIS Models & Input & \# Param. & GPU (GB)& t (s) &Average BA\\
    \midrule
    {LoTeNet (ours)}  & 3D & 52M & $2.1$  &45.3& ${\bf 0.71 \pm 0.09}$ \\
    {Subject-level CNN} & 3D & 1M & $8.8$  &8.7& $0.67 \pm 0.08$\\ 
    {CNN Baseline} & 3D &6.4M& $11.5$  &12.8& $0.64 \pm 0.05$\\ 
    {MLP Baseline} & 3D &78M& $4.5$  &4.1& $0.63 \pm 0.03$\\ 
    \hline
    {Densenet} & 2D &0.2M& $10.5$  &80.1& $0.67 \pm 0.04$\\
    {LoTeNet (ours) } &  2D &0.4M& $0.7$  &81.3& $0.65 \pm 0.03$ \\
    \bottomrule
  \end{tabular}
\end{table}

\section{Discussion}
\label{sec:disc}

In this section, we focus on discussing high level trends observed with experiments reported in Section~\ref{sec:exp}, conceptual connections to other methods, and point out possible directions for future work.

{
\subsection{Comparison with tensor network models}
    The proposed LoTeNet model is related to the tensor network models used for supervised learning tasks, such as the ones in~\citet{stoudenmire2016supervised,efthymiou2019tensornetwork}. In~\citet{stoudenmire2016supervised}, the parameters of the tensor network are learned from data using density matrix renormalization group (DMRG) algorithm~\citep{mcculloch2007density}, whereas in LoTeNet we use automatic differentiation based on the implementation in~\citet{torchmps} similar to~\citet{efthymiou2019tensornetwork} to optimise the parameters of the model in Eq.~\eqref{eq:mps}.}

{
    The proposed LoTeNet model also improves upon the computation cost of approximating linear decisions compared to other tensor network models such as in~\citet{efthymiou2019tensornetwork}. Consider the computation cost of approximating the linear decision in Eq.~\eqref{eq:linModel} for an input image with $N$ pixels of $d$ features with a bond dimension $\beta$. For the tensor network in~\citet{efthymiou2019tensornetwork} it is $\mathcal{O}\left( N \cdot d \cdot \beta^2\right)$. For LoTeNet, the computation cost reduces exponentially with the number of layers used: $\mathcal{O}\left( \frac{N }{k^{2\cdot L}} \cdot L \cdot k^2 \cdot d \cdot \beta^2\right)$. The spatial resolution of the input image is reduced with successive layers in LoTeNet (Figure~\ref{fig:LoTeNet}) captured as the $k^{2\cdot L}$ term in the denominator. The cost of performing MPS operations on patches of size $k\times k$ in $L$ layers is the additional $L\cdot k^2$ term in the numerator. However, for $L \geq 2$ and $k >  1$, the exponential reduction in computation cost with $L$ can translate into considerable computational advantage when processing high resolution medical images. Further, the computation cost of LoTeNet can be reduced without noticeable degradation in performance by sharing MPS in each layer across all patches~\citep{selvan2020multi}.} 

\subsection{On Performance}

\begin{wrapfigure}{r}{0.5\textwidth}
    \vspace{-1cm}
\centering
  {\includegraphics[width=0.95\linewidth]{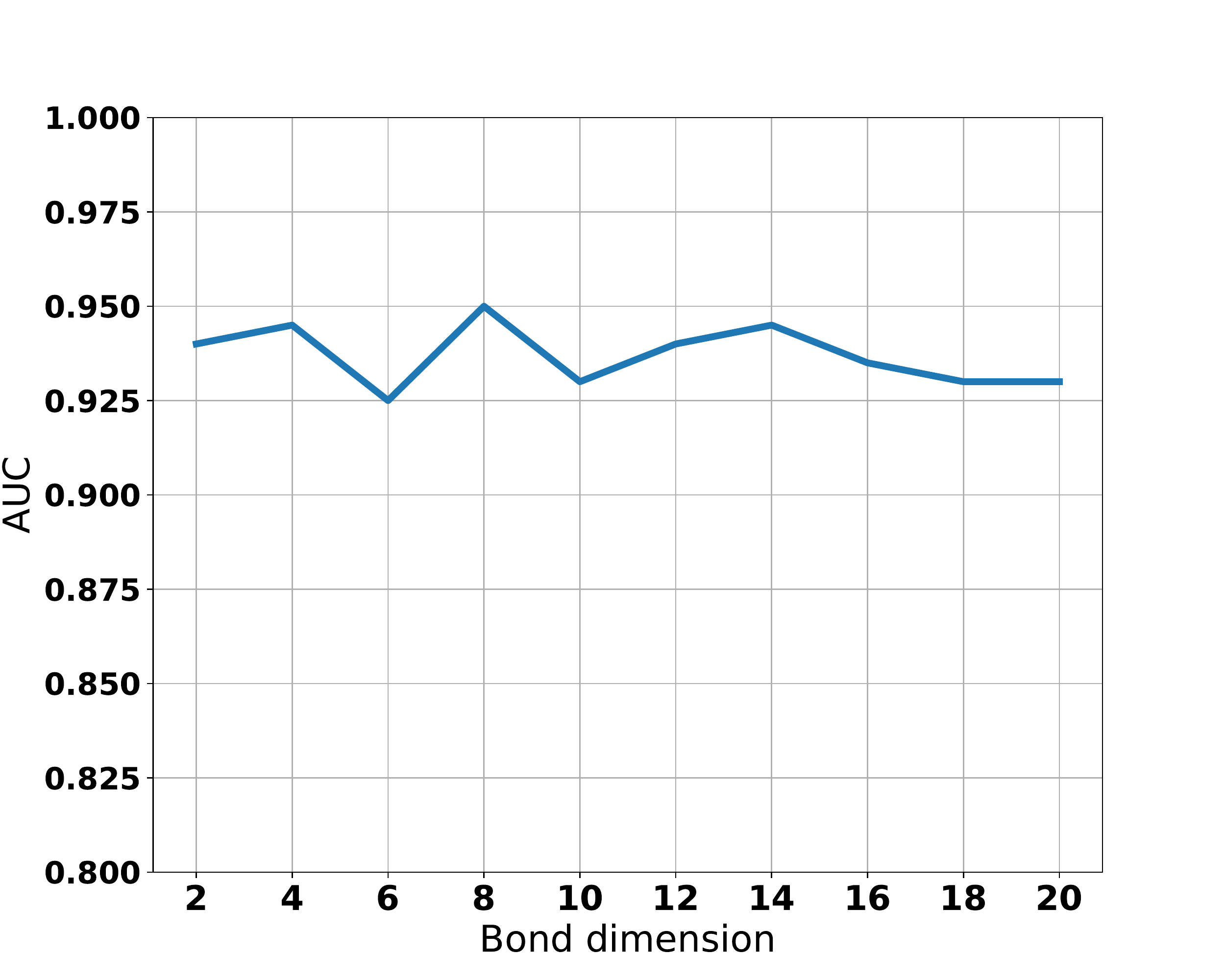}}
  \caption{Influence of varying the bond dimensions, $\beta$ on the LIDC dataset. Note the y-axis is between [0.8-1.0] for better visualisation.}
  \label{fig:bondDim}
\end{wrapfigure}

The performance of  LoTeNet  across the three datasets is on par with, or superior, to the other methods as reported in Tables~\ref{tab:pcam} and~\ref{tab:oasis}. This robustness across tasks is remarkable because the LoTeNet architecture is fixed across the tasks. This behaviour could be attributed to the single model hyperparameter --
bond dimension, $\beta$ -- which is fixed ($\beta=5$) across all experiments. Recollect that the bond dimension controls the quality of MPS approximations of the corresponding operations in high dimensional spaces as described in Section~\ref{sec:mps}. Consistent with the reporting in earlier works~\citep{efthymiou2019tensornetwork}, we also find that the value of $\beta$ after a specific value does not affect the model's performance. 
This is illustrated in Figure~\ref{fig:bondDim}, where we show the sensitivity of the validation set performance on LIDC dataset for different bond dimensions. 
This is a desirable feature as the models can be expected to yield comparable performance when applied to different datasets without entailing dataset specific cost of elaborate architecture search. Additionally, this robustness could also be attributed to more efficient computation of massive number of parameters in LoTeNet resulting in generalised decision rules, as reported in Table~\ref{tab:oasis} where we notice that the number of parameters in LoTeNet are higher than other methods (except MLP baseline).

In Tables~\ref{tab:pcam} and~\ref{tab:oasis}, we report the GPU memory requirement for each of the models. LoTeNet requires only a fraction of the GPU memory utilised by the corresponding baseline methods, including the Densenet~\citep{huang2017densely} or Rotation Eq-CNN models~\citep{veeling2018rotation}. One reason for this drastic reduction in GPU memory utilisation is because tensor networks do not maintain massive intermediate feature maps. This behaviour is similar to feedforward neural networks and unlike CNNs which use a large chunk of GPU memory mainly to store intermediate feature maps~\citep{rhu2016vdnn}. Further, as LoTeNet is based on contracting input data into smaller tensors in a hierarchical manner the memory consumption with successive contracted layers does not increase. This can be an important feature in medical imaging applications as larger images and larger batch sizes can be processed without compromising the quality of the learned decision rules. 


{
\subsection{Choice of local feature maps}
The local feature maps in Equations~\eqref{eq:localRef} and~\eqref{eq:localRef_1} are simple transformations which have been used in the tensor network literature~\citep{stoudenmire2016supervised,efthymiou2019tensornetwork}. More recently, other intensity based local feature maps such as wavelet transforms have been attempted for 1D signal classification in~\citet{reyes2020multi}. While we use the local map in Equation~\eqref{eq:localRef} for 2D data and squeeze operation based local map for 3D data, experiments on other possible local feature maps were also performed. For instance, we experimented with jet based feature maps~\citep{larsen2012jet} that compute spatial gradients at multiple scales, and also an MLP based local feature map acting on individual pixels. Both these local feature maps did not yield any considerable performance improvement and to adhere to existing tensor network literature we used the simple sinusoidal feature maps which also do not have additional hyperparameters.}
    

\subsection{Comparison with Neural Networks}
\begin{figure}[t]
    \centering
    \includegraphics[width=0.75\textwidth]{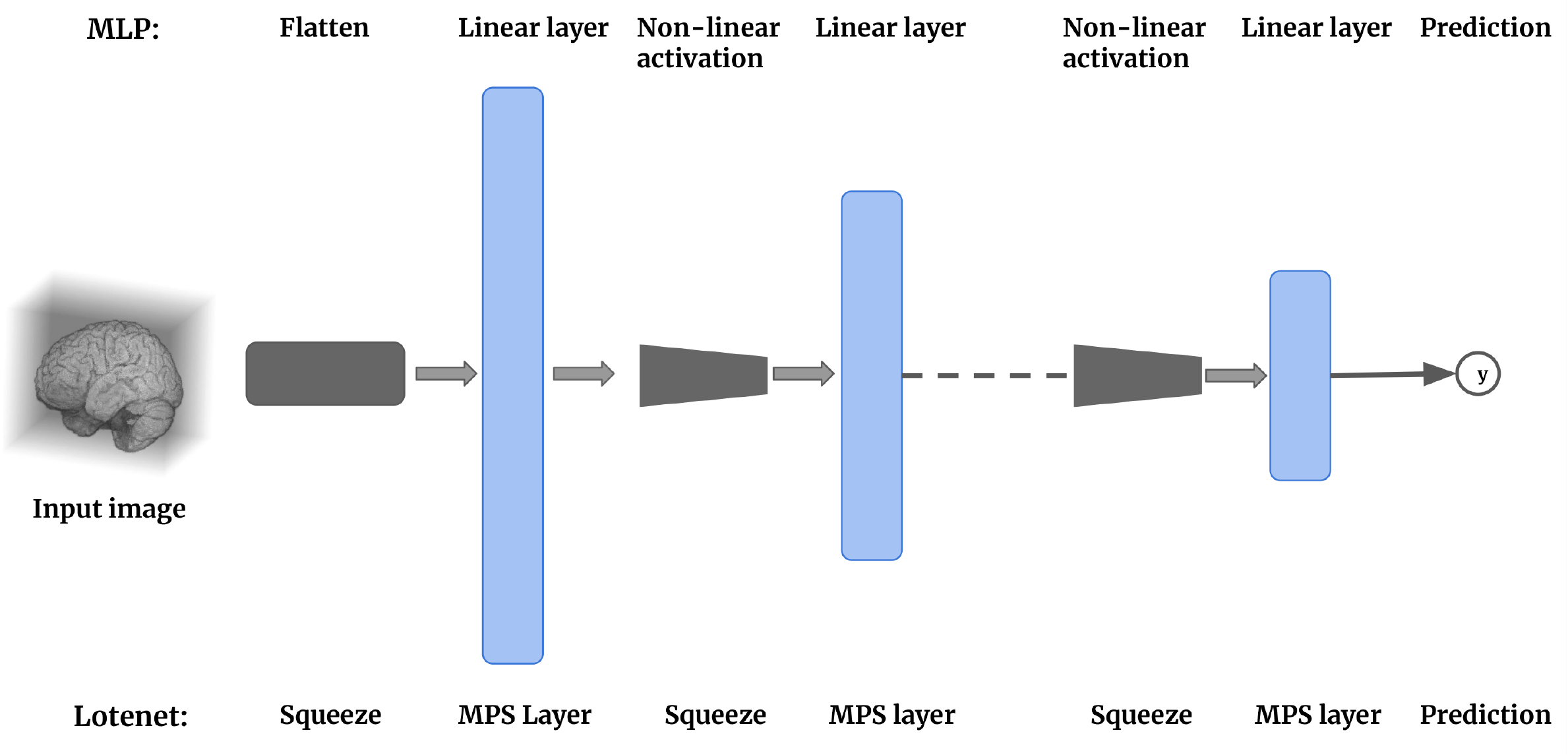}
    \caption{Conceptual comparison of a feedforward neural network such as an MLP with the proposed LoTeNet model. Notice the functional correspondence between linear and MPS layers, and non-linear activation functions and the squeeze operations between the MLP and LoTeNet models, respectively. The input image is flattened to a vector to input to the MLP model, whereas for the LoTeNet model we squeeze small image regions before flattening as described in Section~\ref{sec:squeeze}.}
    \label{fig:modelComp}
\end{figure}

{\bf Feed-forward neural networks:} The class of tensor network models adapted for machine learning tasks~\citep{stoudenmire2016supervised,efthymiou2019tensornetwork} are conceptually most similar to feed-forward neural networks such as the MLP, as both classes of models operate on vector inputs (see Figure~\ref{fig:modelComp}). The difference, however, arises in the type of decision boundaries optimised by each class of model as illustrated in Figure~\ref{fig:decision}. MLP based feed-forward networks obtain non-linear decision boundaries using several layers of neurons and non-linear activation functions (such as sigmoid, ReLU etc.). Tensor network models, including the proposed model, optimise linear decision boundaries in exponentially high dimensional spaces without using any non-linear activation functions. 
\\
{\bf Convolutional neural networks:} LoTeNet uses MPS blocks on image patches and aggregates these tensor representations in a hierarchical manner to learn the decision rule. While this could appear similar to the use of strided convolution kernels in CNNs,
it is indeed closer to feed-forward neural networks. The use of an MPS block per patch perhaps can be interpreted as a form of strided convolution but only in how the local operation is performed on each patch. The primary difference is in the weight sharing; the weights of MPS blocks are not shared across the image. Each $k\times k$ image region has an MPS block acting on it. This is indicated in Figure~\ref{fig:LoTeNet}, where we report the number of MPS blocks used per layer. 

\subsection{Limitations and Future Work}

The computation time reported in Table~\ref{tab:oasis} shows it to be higher for LoTeNet ($45.3$s) compared to the CNN ($12.8$s) or MLP ($4.1$s) baselines as efficient implementations of tensor contractions are not natively supported in frameworks such as PyTorch. There are ongoing efforts to further improve efficient implementations which can accelerate tensor network operations~\citep{fishman2020itensor,novikov2020tensor}. Considering that LoTeNet usually converges between 10-20 epochs, the increase in computation time with the current implementation is not substantial as the model can be trained even on 3D data easily under an hour. 

One possible drawback of having a single hyperparameter $\beta$ controlling the MPS approximations is the lack of granular control of the model capacity. In the 2D OASIS experiments (Table~\ref{tab:oasis}), we further investigated the lower average balanced accuracy score by trying out different $\beta$ values. This experiment revealed that LoTeNet was either under-fitting ($\beta < 4$) or over-fitting ($4 \leq \beta \leq 8$) to the training set. This could be attributed to the fact that the number of parameters grow quadratically with $\beta$ and these discrete jumps make it harder to obtain models of optimal capacity for some tasks.

Another concern with tensor network based models is their tendency to be over parameterised as the number of parameters scale as \{$d\cdot N \cdot \beta^2$\}. The linear dependency on the number of pixels, $N$, for volumetric data results in a model with massive number of parameters at the outset which is further aggravated by the quadratic dependency on $\beta$. While in our experiments, this has not had an adverse effect on the performance, we have observed the model is capable of over-fitting on small training data quite easily. This can be a concern when dealing with datasets with few samples. 

One way of handling these problems would be to introduce weight-sharing between MPS blocks at each layer so that the number of parameters can be reduced and controlled better. A possible future work is to investigate the effect of sharing MPS blocks per layer and study if equivariance to translation can be induced in tensor network based models~\citep{cohen2016group}. 

Finally, the issue of extending LoTeNet type models beyond classification or regression tasks to more diverse medical imaging tasks such as segmentation and registration are open research questions waiting to be explored further. Early work on using tensor networks for 2D segmentation tasks was recently reported in~\citet{selvan2021strided}. 

\vspace{-0.35cm}
\section{Conclusion}
\label{sec:conc}

We have presented a tensor network model for classification of 2D and 3D medical images, of higher spatial resolutions. Using LoTeNet we have shown that aggregating MPS approximations on small image regions in a hierarchical manner retains global structure of the image data. With experiments on three datasets we have shown such models can yield competitive performance on a variety of classification tasks. We have measured the GPU memory requirement of the proposed model and shown it to be a fraction of the memory utilisation of state-of-the-art CNN based models when trained to attain comparable performance.  Finally, we have demonstrated that a single model architecture with $\beta=5$ (tuned on one of the datasets) can perform well on other datasets also, which can be an appealing feature as it reduces dataset specific architecture search.

\vspace{-0.35cm}
\acks{
    Data used in this work were provided in part by OASIS: Cross-Sectional: Principal Investigators: D. Marcus, R, Buckner, J, Csernansky J. Morris; P50 AG05681, P01 AG03991, P01 AG026276, R01 AG021910, P20 MH071616, U24 RR021382. We thank Eric Kim for permitting us to modify and use Figure~\ref{fig:decision}. The authors also thank all reviewers and editors for their insightful feedback.} 

\vspace{-0.35cm}
\ethics{The work follows appropriate ethical standards in conducting research and writing the manuscript. All data used in this project are from open repositories.}

\vspace{-0.35cm}
\coi{We declare we do not have conflicts of interest.}
\bibstyle{plain}
\setlength{\bibsep}{4pt}
\bibliography{selvan20}





\end{document}